\RequirePackage{amsmath}
\documentclass[runningheads]{llncs}

\usepackage[T1]{fontenc}
\usepackage{amsmath,amssymb,amsfonts,bm}
\usepackage{cite}
\usepackage{floatrow}
\usepackage{ulem}
\usepackage{algorithmic}
\usepackage{graphicx}
\usepackage{textcomp}
\usepackage{xcolor}
\usepackage{booktabs}
\usepackage{hyperref}
\usepackage{cleveref}
\usepackage{url}
\usepackage[english]{babel}
\usepackage{caption}
\usepackage{subcaption}
\usepackage[misc]{ifsym}

\def\codeurl{https://professor-x.de/dynabench}

\usepackage{color}

\newfloatcommand{capbtabbox}{table}[][\FBwidth]
\begin{document}
\title{DynaBench: A benchmark dataset for learning dynamical systems from low-resolution data.}
\titlerunning{DynaBench}
%
\author{Andrzej Dulny\orcidID{0009-0002-2990-9480}~\Letter \and
Andreas Hotho\orcidID{0000-0002-0483-5772} \and
Anna Krause\orcidID{0000-0003-1924-9183}}
\authorrunning{A. Dulny et al.}
\tocauthor{Andrzej~Dulny}
\toctitle{DynaBench:~A benchmark dataset for learning dynamical systems from low-resolution data.}

%
\institute{University of Würzburg, Germany \\
\email{\{dulny,andreas.hotho,anna.krause\}@uni-wuerzburg.de}}
\maketitle              
\begin{abstract}

Previous work on learning physical systems from data has focused on high-resolution grid-structured measurements. 
However, real-world knowledge of such systems (e.g. weather data) relies on sparsely scattered measuring stations.
In this paper, we introduce a novel simulated benchmark dataset, DynaBench, for learning dynamical systems directly from sparsely scattered data without prior knowledge of the equations. 
The dataset focuses on predicting the evolution of a dynamical system from low-resolution, unstructured measurements. 
We simulate six different partial differential equations covering a variety of physical systems commonly used in the literature and evaluate several machine learning models, including traditional graph neural networks and point cloud processing models, with the task of predicting the evolution of the system.
The proposed benchmark dataset is expected to advance the state of art as an out-of-the-box easy-to-use tool for evaluating models in a setting where only unstructured low-resolution observations are available.
The benchmark is available at~\url{\codeurl}.



\end{abstract}
\section{Introduction}
\label{lab:introduction}
Dynamical systems, which are systems described by partial differential equations (PDEs), are ubiquitous in the natural world and play a crucial role in many areas of science and engineering.
They are used in a variety of applications, including weather prediction~\cite{bauer_quiet_2015}, climate modeling~\cite{Cullen1997}, fluid dynamics~\cite{Kleinstreuer2010}, electromagnetic field simulations~\cite{Sheikholeslami2018} and many more. 
Traditionally, these systems are simulated by numerically solving a set of PDEs that are theorized to describe the behavior of the system based on physical knowledge.
An accurate modelling technique is crucial for ensuring accurate predictions and simulations in these applications.
However, the equations used are often just an approximation of a much more complex reality, either due to the sheer complexity of a more accurate model which would be computationally infeasible or because the true equations are not known~\cite{McGuffie2001}.

In recent years, several models have been proposed in the deep learning community, which address the problem of simulating physical systems by learning to predict dynamical systems directly from data, without knowing the equations a priori~\cite{Iakovlev2021a,Li2021,Berg2019,Praditia2021,Dulny2022}.
These types of approaches have a distinct advantage over classical numerical simulations, as they do not require estimating the parameters of the equations, such as the permeability of a medium or the propagation speed of a wave.
To ensure that the proposed models and architectures perform and generalize well and to be able to draw a fair comparison between them, it is necessary to compare them in a common experimental setting.
As there are very few real-world datasets readily available for this purpose, it is common practice to employ simulated data as a simplified but easy-to-use and available alternative to evaluate novel machine learning methods~\cite{Karlbauer2019,Dulny2022,Ayed2019,Takamoto2022,Anandkumar2022}.

\begin{figure}[h]
\includegraphics[width=1\textwidth]{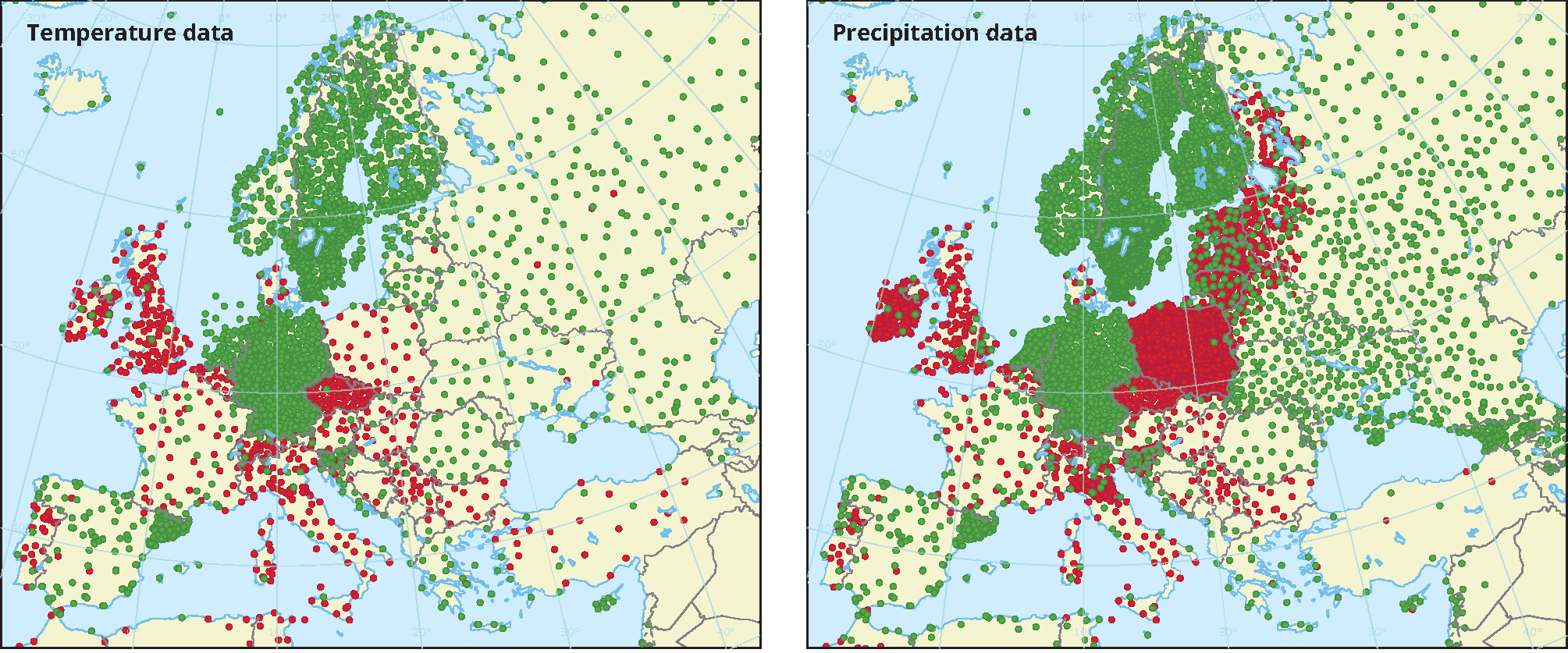}
\caption{Map of weather stations within the European Climate Assessment and Datasets (ECA\&D) monitoring network for temperature and precipitation data~\cite{Haylock2008}. Monitoring stations are not located on a grid but instead strategically placed based on a variety of factors such as topography, accessibility, and weather patterns.}
\label{fig:monitoring_stations}
\end{figure}

While some progress has been made towards creating a standardized benchmark~\cite{Huang2021, Otness2022, Takamoto2022} dataset of physical simulations, the previous work in this area mainly focuses on the task of reconstructing the forward operator of the numerical solver, for which the full computed solution on a high-resolution grid of the differential equation is needed as training data.
This makes it difficult to assess the applicability of any approach evaluated this way on real data, where measurements are typically neither high resolution nor grid-based, but instead rely on a sparse network of measuring stations (cf.~\Cref{fig:monitoring_stations}).

To achieve greater fidelity to real-world conditions, we propose a novel benchmark dataset, DynaBench, that focuses on the challenging task of predicting the evolution of a dynamical system using a limited number of measurements that are arbitrarily distributed within the simulation domain.
This more closely resembles a real-world setting and allows for a more accurate assessment of the applicability of different models to real-world data.
The benchmark consists of simulations of six physical systems with different properties that are commonly used as synthetic data for learning dynamical systems.
The simulations have been generated using a numerical solver.
Our aim is not to cover all possible physical systems, parameters, and equations but rather to provide a good starting point to develop and compare machine learning models suited for this task.
The selection we propose is a combination of typical equations used to evaluate deep learning models and equations with different properties (such as order of derivatives and number of variables) that complement them. 

In addition, we present a detailed evaluation of various comparison models capable of learning functions on arbitrary geometries, including graph neural networks~\cite{Iakovlev2021a,kipf2017semisupervised,veličković2018graph,Gilmer2017}, point cloud neural networks~\cite{Qi_2017_CVPR,Zhao2021,Shi2020}, and continuous convolution models~\cite{wang2018contconv,Thomas2019}.
Our objective is to provide a set of strong baselines for further research, and thus facilitate the development and testing of new machine learning methods for predicting physical systems from unstructured low-resolution data.
Our results show that the selected models are capable of providing accurate short-term predictions, but long-term forecasting remains an open challenge.

With the release of DynaBench, we hope to provide a valuable resource for the machine learning community, which will facilitate research and thus advance the state-of-the-art in learning dynamical systems from data on unstructured low-resolution observations.

The main contributions of our work can be summarized as follows.
\begin{enumerate}
    \item We propose a new benchmark dataset for learning dynamical systems from data under the assumption that measurements are sparse and not structured on a grid.

    \item We generate the dataset by simulating several differential equation systems typically used for the task of learning dynamical systems.
    
    \item We thoroughly evaluate several models capable of learning functions on arbitrary geometries on the DynaBench dataset, including both graph neural networks and point-cloud processing models.
    
    \item We release both the dataset and the code for evaluating all models, to facilitate further research in this field~\footnote{The benchmark is available at \url{\codeurl}}.
\end{enumerate}

\section{Related Work}
\label{sec:related_work}
Several approaches for learning dynamical systems from grid data have been proposed in recent years, but they lack comparability as different sets of equations and simulation parameters are used.
Ayed et al.~\cite{Ayed2019} propose a hidden-state neural solver-based model and use a system of shallow water equations and an Euler fluid simulation to evaluate it.
Long et al.~\cite{Long2019} evaluate their numeric-symbolic hybrid model on the Burgers' equation, diffusion equation and convection-diffusion equation with a reactive source.
Dulny et al.~\cite{Dulny2022} evaluate their neuralPDE Model based on neural solvers on several PDE systems, including advection-diffusion, Burgers' and wave equations.
Li et al.~\cite{Li2021} propose a resolution invariant method based on the fourier transformation and test it on Burgers' equation, simplified Navier-Stokes system and steady-state darcy flow.

Similarly, authors proposing models for unstructured data (i.e. measurements not on a grid) also do not evaluate their models on a common set of systems.
Karlbauer et al.~\cite{Karlbauer2019} propose a graph-based recurrent model (Distana) to learn spatio-temporal processes and evaluate it on the wave propagation equation.
Iakovlev et al.~\cite{Iakovlev2021a} use an advection-diffusion problem, as well as the heat equation and Burger's equation, to evaluate their graph message passing approach.
Another approach proposed by Li et al.~\cite{Li2020}, the multipole graph neural operator, is evaluated on the steady state darcy flow, as well as the viscous variant of the Burgers' equation.

Recently, some progress has been made towards creating a standardized benchmark for learning PDEs from data.
Huang et al.~\cite{Huang2021} proposed a dataset containing simulations of incompresible Navier-Stokes equations for fluid dynamics.
While the main audience of the dataset is not the machine learning community, as its central purpose is to compare different discretization and solving schemes, the data could in theory still be used to train different models for learning the solutions from data.
However, it remains limited in the choice of equations, as it only uses the Navier-Stokes equations, and furthermore is not suited for evaluating models in a low-resolution regime.
Otness et al.~\cite{Otness2022} propose a benchmark specifically aimed at learning to simulate physical systems from data.
However, the simulations are discrete systems (spring systems) rather than continuous spatiotemporal processes defined by partial differential equations.
For this reason they cannot be used for the intended purpose of learning continuous systems from low-resolution measurements.

Takamoto et al.~\cite{Takamoto2022} propose a very extensive benchmark of eleven different equation systems called PDEBench, including fluid simulations, advection and diffusion equations, Burgers' equation and more.
The authors also provide extensive experiments and evaluations for a variety of models.
The benchmark is well suited for learning in a high-resolution framework, where the whole discretized grid used during numerical solving is also used for training the models.
However, the selection of equations consisting mainly of fluid simulations is unsuitable for low-resolution predictions, as such systems show turbulent and chaotic behavior~\cite{foias_manley_rosa_temam_2001,Deissler1986} and therefore require a high-resolution discretization.
As such PDEBench is neither suited nor easily usable in a low-resolution regime, where only limited number of scattered observation are available.

\section{Dataset}
In this section we describe the overall structure of the datasets, which equations were included in the benchmark, how the simulations were executed, and what postprocessing steps were performed.

\subsection{Setting}
A PDE is a equation in which an unknown function is to be found, based on the relations between itself and its partial derivatives in time and space.
It can be summarized in the form:
\begin{equation}
    F(u, \frac{\partial u}{\partial t}, \frac{\partial u}{\partial x}, \frac{\partial u}{\partial y}, ...)=0
\end{equation}
As mentioned in~\Cref{lab:introduction} such equations can be used to model a variety of physical systems, by solving a previously known equation system using a measured initial state.
In the context of scientific machine learning, a typically researched task is to reconstruct the parameters of the equation (i. e. the function $F$) from data obtained from a mixture of exact measurements and simulations.
Reconstructing the differential equations requires high-resolution data (both in time and space), which is unavailable in a real world setting~\cite{Dulny2022}.
Our benchmark is focused on a different task, namely learning to predict the evolution of a dynamical system from data, under the assumption that only low-resolution measurements are available.
Formally, a PDE solver seeks to approximate the true solution
\begin{equation*}
    u\colon\Omega\times T\longrightarrow \mathbb{R}
\end{equation*}
by some approximate
\begin{equation*}
    \hat{u}_{h}\colon\hat{\Omega}_{h}\times \hat{T}_{h}\longrightarrow \mathbb{R},
\end{equation*}
where $\hat{\Omega}_{h}$ is a high-resolution discretization of the solution domain $\Omega\subseteq\mathbb{R}^n$ (typically a grid) and $\hat{T}_{h}$ is a high-resolution time discretization of $T\subseteq\mathbb{R}$ (typically $\hat{T}_{h} = \{t_k^{(h)}, k\in\mathbb{N}\}$ for $t_k^{(h)} := t_0 + k\Delta_h t$ and some small $\Delta_h t > 0)$.

For our task we assume that only low-resolution observations $\hat{u}_l$ at measurement locations $\hat{\Omega}_{l}$ of the physical process $u$ are available (i.e. $|\hat{\Omega}_{l}|\ll|\hat{\Omega}_{h}|$), and the temporal resolution $\hat{T}_{l} = \{t_k^{(l)}, k\in\mathbb{N}\}$ for $t_k^{(l)} := t_0 + k\Delta_l t$ of the measurements is also low ($|\Delta_l t| \ll \Delta_h t$).
The task is then to predict the evolution of the system $\hat{u}_l(\hat{\Omega}_l, t_{k+1}^{(l)}), \hat{u}_l(\hat{\Omega_l}, t_{k+2}^{(l)}), \ldots, \hat{u}_l(\hat{\Omega}_l, t_{k+R}^{(l)})$, from the past observations $\hat{u}_l(\hat{\Omega}_l, t_{k-H}^{(l)}), \ldots, \hat{u}_l(\hat{\Omega}_l, t_{k-1}^{(l)}), \hat{u}_l(\hat{\Omega}_l, t_{k}^{(l)})$.
\subsection{Equations}
Overall we curated a set of six different PDE equation systems, typically used in the context of learning dynamical systems from data, with various properties as summarized in~\Cref{tab:equation_summary}.
In the following we shortly describe each equation in more detail.

\begin{table}[ht]
    \caption{Summary of the PDE systems used in our benchmark dataset}
    \centering
    \renewcommand{\arraystretch}{1.1}
    \begin{tabular}{lccc}
        \toprule
         Equation & Components & Time Order & Spatial Order \\
         \midrule
         Advection & 1 & 1 & 1 \\
         Burgers & 2 & 1 & 2 \\
         Gas Dynamics & 4 & 1 & 2 \\
         Kuramoto-Sivashinsky & 1 & 1 & 4 \\
         Reaction-Diffusion & 2 & 1 & 2 \\
         Wave & 1 & 2 & 2 \\
         \bottomrule
    \end{tabular}
    \label{tab:equation_summary}
\end{table}

\vspace{0.15cm}\noindent
\textbf{Advection}
The advection equation
\begin{equation}
    \frac{\partial u}{\partial t} = - \nabla\cdot (\mathbf{c}u)
\end{equation}
describes the displacement of a quantity described by a scalar field $u$ in a medium moving with the constant velocity $\mathbf{c}$.
It is a widely used benchmark equation due to its simplicity and straightforward dynamics~\cite{Dulny2022, Long2019}

\vspace{0.15cm}\noindent
\textbf{Burgers' Equation} The Burgers' equation 
\begin{equation}
    \frac{\partial \mathbf{u}}{\partial t} = R(\nu \nabla ^2 \mathbf{u} - \mathbf{u}\cdot\nabla \mathbf{u})
\end{equation}
is a non-linear second order PDE with respect to spatial derivatives

The equation describes the speed $u$ of a fluid in space and time with $\nu$ representing the fluid's viscosity and $R$ describing the rate of the simulation.
It is one of the most often used equations in the context of deep learning for dynamical systems~\cite{Dulny2022,Takamoto2022,Iakovlev2021a,Li2020}.

\vspace{0.15cm}\noindent
\textbf{Gas Dynamics}
In gas dynamics, the system of coupled non-linear PDEs
\begin{equation}
    \begin{split}
        &\frac{\partial \rho}{\partial t} = -\mathbf{v}\cdot\nabla\rho - \rho\nabla\cdot\mathbf{v}\\
        &\frac{\partial T}{\partial t} = -\mathbf{v}\cdot\nabla T - \gamma T\nabla\cdot\mathbf{v} + \gamma\frac{Mk}{\rho}\nabla^2 T\\
        &\frac{\partial \mathbf{v}}{\partial t} = - \mathbf{v}\cdot \nabla\mathbf{v} - \frac{\nabla P}{\rho} + \frac{\mu}{\rho}\nabla(\nabla\mathbf{v})\\
    \end{split}    
\end{equation}
describes the evolution of temperature $T$, density $\rho$, pressure $P$ and velocity $\mathbf{v}$ in a gaseous medium.
The equations are derived from the physical laws of mass conservation, conservation of energy, and Newton's second law~\cite{Anderson1995}.
The parameters specify the physical properties of the system, $\gamma$ being the heat capacity ratio, $M$ the mass of a molecule of gas, and $\mu$ the coefficient of viscosity.
This equation can be seen as a simplified weather system.

\vspace{0.15cm}\noindent
\textbf{Kuramoto-Sivashinsky}
The Kuramoto-Sivashinsky equation
\begin{equation}
\begin{split}
    \frac{\partial u}{\partial t} = - \frac{1}{2}|\nabla u|^2 - \nabla^2u - \nabla^4u
\end{split}
\end{equation}
describes a model of the diffusive–thermal instabilities in a laminar flame front.
Solutions of the Kuramoto–Sivashinsky equation possess rich dynamical characteristics~\cite{Cvitanovic2010} with solutions potentially including equilibria, relative equilibria, chaotic oscillations and travelling waves.

\vspace{0.15cm}\noindent
\textbf{Reaction-Diffusion}
The Reaction-Diffusion system
\begin{equation}
\begin{split}
    &\frac{\partial u}{\partial t} = D_u\nabla^2 u + a_u(u - u^3 - k - v)\\
    &\frac{\partial v}{\partial t} = D_v\nabla^2 v + a_v(u-v)
\end{split}
\end{equation}
describes the joint concentration distribution of a two component chemical reaction, where one of the components stimulates the reaction and the other inhibits it.
The parameters $D_u$ and $D_v$ describe the diffusion speed of the activator and inhibitor respectively, $k$ is the activation threshold, while $a_u$ and $a_v$ describe the reaction speed of the two components.
The equation has applicability in describing biological pattern formation and forms rich and chaotic systems~\cite{Takamoto2022,FitzHugh1961}.

\vspace{0.15cm}\noindent
\textbf{Wave}
The wave equation
\begin{equation}
    \frac{\partial^2 u}{\partial t^2} = \omega^2 \nabla ^2 u
\end{equation}
describes the propagation of a wave in a homogeneous medium (e.g. water surface) where $u$ describes the distance from equilibrium and $\omega$ represents the material-dependent speed of propagation.
It is a linear, second-order PDE that has been widely used in scientific machine learning~\cite{Dulny2022,Karlbauer2019,Moseley2020,Otte2020,Karlbauer2020}.

\subsection{Simulation Parameters}
The machine learning task for which our benchmark has been designed, is to learn predictions from observations of a physical system. 
The system is assumed to evolve according to a set of fixed physical laws
that are have constant parameters such as thermal conductivity, diffusion coefficients etc.
To create simulations of such systems, we specify the constant parameters with which the selected equations are solved, as shown~\Cref{tab:equation_parameters}.
The parameters have been chosen to ensure a good balance between the complexity of the system and the numerical stability of the simulations.

\begin{table}[ht]
    \caption{Equation parameters used for the simulations}
    \centering
    \renewcommand{\arraystretch}{1.1}
    \begin{tabular}{lc}
        \toprule
         Equation & Parameters \\
         \midrule
         Advection & $c_x=1$, $c_y=1$ \\
         Burgers & $\nu = 0.5$, $R=25$\\
         Gas Dynamics & $\mu = 0.01$, $k=0.1$, $\gamma = 1$, $M=1$ \\
         Kuramoto-Sivashinsky & -\\
         Reaction-Diffusion & $D_u=0.1$, $D_v=0.001$, $k = 0.005$, $a_u=1$, $a_v=1$ \\
         Wave & $\omega=1$ \\
         \bottomrule
    \end{tabular}
    \label{tab:equation_parameters}
\end{table}

\begin{figure}[ht]
\begin{floatrow}
\capbtabbox{%
    \begin{tabular}{lc@{\hskip 0.1in}c}
        \toprule
         Equation & Field & Initial Cond. \\
         \midrule
         Advection & $u$ & gaussian \\
         Burgers & $u$ & gaussian \\
           & $v$ & gaussian \\
         Gas Dynamics & $\rho$ & gaussian\\
          & $T$ & gaussian \\
          & $v_x$ & zero \\
          & $v_y$ & zero \\
         Kuramoto-Sivashinsky & $u$ & noise (u)\\
         Reaction-Diffusion & $u$ & noise (n) \\
          & $v$ & noise (n) \\
         Wave & $u$ & gaussian \\
          & $\frac{\partial u}{\partial t}$ & zero\\
         \bottomrule
    \end{tabular}
}{%
  \caption{Initial conditions used for each system}%
  \label{tab:initial_conditions}%
}
\ffigbox[0.4\textwidth]{%
  \caption{Example of a gaussian initial condition as defined in ~\Cref{eq:initial_condition}}%
}{%
  \includegraphics[width=0.4\textwidth]{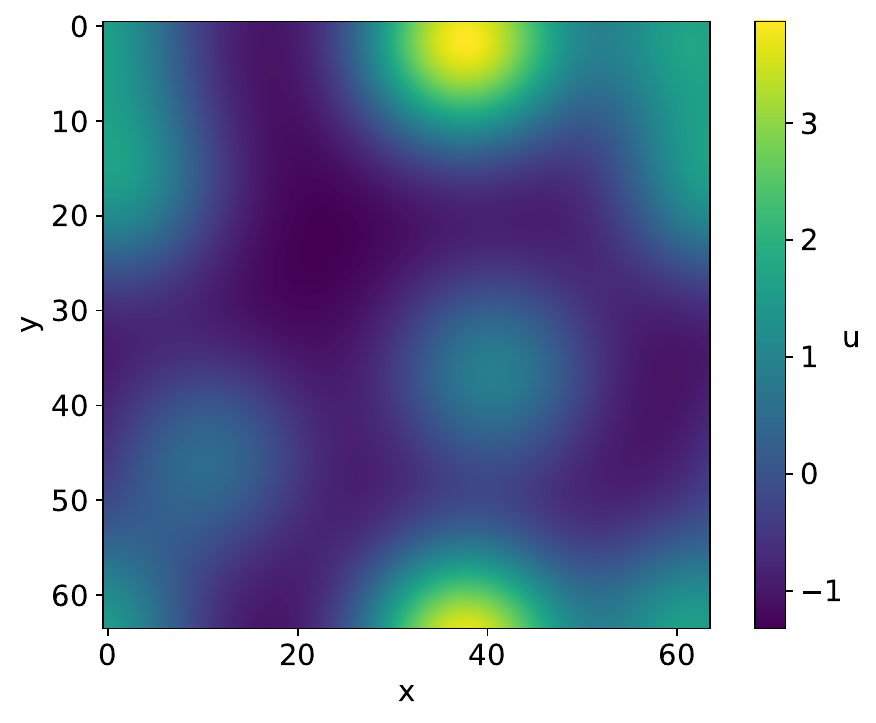}
}
\end{floatrow}
\label{fig:initial_conditions}
\end{figure}

The spatial domain of the simulation is set to $\Omega=[0, 1]\times[0, 1]$ and the temporal domain to $T=[0, 200]$.
We initialize the state of each system using zeros, uniform (u) or normally (n) distributed noise, or a sum of Gaussian curves, individually for each field, similar to what has been used in related work~\cite{Karlbauer2019, Dulny2022, Takamoto2022}.
The exact specification of which initial condition is used for each individual variable is summarized in~\Cref{tab:initial_conditions}.
The sum of Gaussian curves has been calculated in the following manner:
\begin{equation}
\label{eq:initial_condition}
    I(x, y) = \sum^{K}_{i=1}A_ie^{-\frac{(x-\mu_{ix})^2+(y-\mu_{iy})^2}{\sigma^2}}
\end{equation}

The positions $(\mu_{ix}, \mu_{iy})$ of each component $i$ are sampled uniformly from the simulation domain $\Omega$, while their contributions $A_i$ are sampled uniformly from the interval $[-1, 1]$.
The fixed parameters $K$ and $\sigma$ are set to $5$ and $0.15$ respectively.

To run the simulations, the domain $\Omega$ is discretized as a $64\times 64$ grid, which yields a cell size of $\Delta x = \Delta y = 0.0156$.
The equations are solved using the method of lines as numerical scheme ~\cite{Dulny2022}.
We use the Explicit Runge-Kutta method of order 5(4)~\cite{Dormand1980} as the numerical integrator.

\subsection{Postprocessing}
\label{sec:postprocessing}

The simulation is saved with a temporal resolution of $\Delta t = 1$, producing exactly $201$ observations per simulation.
As some of the equations produce non-stationary physical processes, we normalize the data to ensure that range of values remains similar across different equations, simulations and times.
Finally, we sample measurements to form the non-grid observation domain, by selecting uniformly $K$ points from the simulation domain $\Omega$ and bilinearily interpolate the values from the grid measurements.

\subsection{Data availability}
\label{sec:data_availablity}
In total we generate 7000 different simulations for each equation, divided into 5000 training simulations and 1000 validation and test simulations each.
For each simulation, we use a different initial seed to sample the initial condition.
The benchmark is available in three different resolutions, where either $K=225$, $K=484$, or $K=900$ measurements are recorded.
Additionally we provide a low-resolution variant of the simulation measured on a grid with the same number of points in total - $15\times 15$, $22\times 22$, $30\times 30$.

The full dataset (including the original high-resolution simulations) can be downloaded at~\url{\codeurl}. 
Alternatively the same data can be generated from scratch using the provided source code and predefined seeds.
Additionally more data can be generated

\section{Experiments}
In this section we describe a selection of experiments that we performed on the DynaBench dataset.

\subsection{Models}
In the following, we briefly describe the models used during the experiments.
We select several graph neural network and point cloud network baselines as a comparison for available state-of-the-art architectures proposed for learning dynamical systems from scattered measurements - graph kernel networks and graph PDE networks.
We do not include Distana~\cite{Karlbauer2019} and Multipole Graph Operator~\cite{Li2020} (cf.~\Cref{sec:related_work}) as there is no code available for the former and the latter requires measurements obtained at different resolution levels and is unsuitable for our setting.

Additionally, to better understand how the change of structure affects the accuracy of the predictions, we evaluate three models that work on grid data trained on a version of the dataset using the same number of measurements but aligned on a grid, as described in~\Cref{sec:data_availablity}.
These include two variants of a simple convolutional neural network~\cite{LeCun1999} - with and without residual connections~\cite{He2016} and neuralPDE, a model specifically designed to learn dynamical systems from gridded data~\cite{Dulny2022}.

Finally, we use the persistence baseline as a reference point for all deep learning models.

\vspace{0.15cm}\noindent
\textbf{PointGNN} is a graph neural network proposed by~\cite{Shi2020} to solve the task of object detection in a LiDAR point cloud.
It uses MLP-based feature aggregation within a local neighborhood with an additional perturbation mechanism to offset the coordinates of the neighboring points.
This increases the translation invariance of the calculated filters with respect to the center vertex coordinates.

\vspace{0.15cm}\noindent
\textbf{Point Transformer} (Point TF) is a model originally proposed by Zhao et al.~\cite{Zhao2021} for object classification and segmentation on 3D point clouds.
It uses self-attention, similar to transformer networks, to process features within a spatially local neighborhood.
We modify the original segmentation architecture to use 2D point coordinates where the physical system has been measured.

\vspace{0.15cm}\noindent
\textbf{Feature-Steered Graph Convolutions} (FeaStNet) is a graph convolution operator developed by Verma at al.~\cite{verma2018feastnet} for 3D object analysis.
It uses the node features from the preceding layer to determine the correspondence between filter weights and nodes in a local neighborhood.
Thus it is able to adjust the filters dynamically based on the final prediction task.

\vspace{0.15cm}\noindent
\textbf{Graph Convolution Network} (GCN) proposed by Kipf et al.~\cite{kipf2017semisupervised} is a simple generalization of convolutions to graph structures where no ordering of the neighbors exists.
It uses a first-order approximation of spectral graph convolutions to aggregate features from neighboring nodes.

\vspace{0.15cm}\noindent
\textbf{Graph Attention Network} (GAT) proposed by Veličković et al.~\cite{veličković2018graph} incorporates an attention mechanism into convolutions on graphs used as weights for aggregating the features from neighboring nodes in each layer.
The attention mechanism is able to (implicitly) assign different weights to different nodes in a neighborhood.

\vspace{0.15cm}\noindent
\textbf{Graph Kernel Network} (KernelNN) is a deep learning approach proposed by Anandkumar et al.~\cite{anandkumar2019neural} for learning a mapping between two infinite-dimensional spaces.
It uses kernel integration with a learnable Nyström kernel as an approximation of the true neural operator.
In the original experiments Anandkumar et al. use a high-resolution grid on which the simulation is computed, but the model itself can be applied to non-grid measurements.

\vspace{0.15cm}\noindent
\textbf{Graph PDE Networks} (GraphPDE) proposed by Iakovlev et al.~\cite{Iakovlev2021a} use the neural network to parameterize the dynamics (rate of change) of the system rather than making predictions directly.
Similar approaches have been proposed for grid data~\cite{Dulny2022,Ayed2019}, outperforming classical architectures for this type of task.
All of these approaches, including graph PDE networks, use the parameterization learned by message passing graph neural networks together with an differentiable ODE solver to obtain predictions.

\vspace{0.15cm}\noindent
\textbf{CNN} originally developed by LeCun et al.~\cite{LeCun1999} uses learnable convolutional filters to enforce translation invariance of the learned mapping with respect to the input position.
While it was originally proposed for computer vision tasks it has since been used in the context of learning to predict dynamical systems from data.
In our experiments we include a simple architecture with several stacked CNN layers, as well as ResNet variant with residual connections~\cite{He2016}.

\vspace{0.15cm}\noindent
\textbf{NeuralPDE} is a model proposed by Dulny et al.~\cite{Dulny2022} combing a convolutional neural network used to parametrize the dynamics (rate of change) of a physical system with differentiable ODE solvers to calculate predictions.
The authors use convolutional layers to approximate partial differential operators, as they directly translate into a discretization using finite differences.
This type of architecture has been shown to perform exceptionally well on a variety of physical data.

\vspace{0.15cm}\noindent
\textbf{Persistence} describes the baseline obtained by applying the rule ``today's weath-er is tomorrow's weather''. 
It suggests the last known input as the prediction of the next state.
Any forecasting model should be able to outperform this baseline, to be counted as useful.
The persistence baseline is a common method used in machine learning for time series forecasting tasks.

\subsection{Setup}
We trained and evaluated all selected models on the DynaBench dataset using 7000 simulations for each equation as training data, and 1000 for validation and testing each.
The input for the models is a $H$-step lookback of the system state (the previous $H$ states) measured at $K$ locations that we merge along the feature dimension.
Specifically, for an physical system describing $D$ variables, the resulting input has the dimension $H\times D$.

We train all models on predicting the next step of the simulation by minimizing the mean squared error (MSE):
\begin{equation}
    \underset{\phi}{\text{min }} \mathbb{E}\big[m_{\phi}(X_t\mathbin\Vert X_{t-1}\mathbin\Vert\ldots\mathbin\Vert X_{t-H+1}) - X_{t+1}\big]^2
\end{equation}
Where $X_{t+1}, X_{t}, X_{t-1},\ldots$ describes the state of the physical system at times $t+1,t, t-1,\ldots$; $m_{\phi}$ is the neural network model with learnable parameters $\phi$; $H$ is the lookback history; and $\mathbin\Vert$ denotes the concatenation operator.

For evaluating the models we rollout $R$ predictions steps in a closed-loop setting where the predictions of previous states are used as input for predicting the new state.
Specifically:
\begin{equation}
\begin{split}
    \hat{X}_{t+1} =& m_{\phi}(X_t\mathbin\Vert X_{t-1}\mathbin\Vert\ldots\mathbin\Vert X_{t-H+1})\\
    \hat{X}_{t+2} =& m_{\phi}(\hat{X}_{t+1}\mathbin\Vert X_{t}\mathbin\Vert\ldots\mathbin\Vert X_{t-H+2})\\
    \hat{X}_{t+3} =& m_{\phi}(\hat{X}_{t+2}\mathbin\Vert \hat{X}_{t+1}\mathbin\Vert\ldots\mathbin\Vert X_{t-H+3})\\
       &\vdots\\
    \hat{X}_{t+R} =& m_{\phi}(\hat{X}_{t+R-1}\mathbin\Vert \hat{X}_{t+R-2}\mathbin\Vert\ldots\mathbin\Vert \hat{X}_{t-H+R})
\end{split}
\end{equation}

In our experiments we use $H=8$, $K=900$ and $R = 16$.

\subsection{Results}
\Cref{tab:results_single} shows the results of our experiments for single-step predictions on the test simulations.
Our results show that non-grid models, such as kernel-based neural networks and graph-based neural networks, can perform similarly to grid-based models for short-term (1-step) predictions. 
Among the models trained on unstructured data, the PointGNN and Point Transformer show the best performance.

\begin{table}[ht]
    \caption{MSE after 1 prediction step. The best perfoming model for each equation has been \underline{underlined}. Additionally, the best non-grid model has been \uwave{underwaved}. A = Advection, B = Burgers', GD = Gas Dynamics, KS = Kuramoto-Sivashinsky, RD = Reaction-Diffusion, W = Wave}
    \centering
    \begin{tabular}{l@{\hskip 0.25cm}l@{\hskip 0.25cm}l@{\hskip 0.25cm}l@{\hskip 0.25cm}l@{\hskip 0.25cm}l@{\hskip 0.25cm}l}
    \toprule
    model &  A &  B &  GD &  KS &  RD &     W \\
    \midrule
    FeaSt             &   $1.30\cdot 10^{-4}$ & $1.16\cdot 10^{-2}$ &      $1.62\cdot 10^{-2}$ &              $1.18\cdot 10^{-2}$ &            $4.89\cdot 10^{-4}$ & $5.23\cdot 10^{-3}$ \\
    GAT               &   $9.60\cdot 10^{-3}$ & $4.40\cdot 10^{-2}$ &      $3.75\cdot 10^{-2}$ &              $6.67\cdot 10^{-2}$ &            $9.15\cdot 10^{-3}$ & $1.51\cdot 10^{-2}$ \\
    GCN               &   $2.64\cdot 10^{-2}$ & $1.39\cdot 10^{-1}$ &      $8.43\cdot 10^{-2}$ &              $4.37\cdot 10^{-1}$ &            $1.65\cdot 10^{-1}$ & $3.82\cdot 10^{-2}$ \\
    GraphPDE          &   $1.37\cdot 10^{-4}$ & $1.07\cdot 10^{-2}$ &      $1.95\cdot 10^{-2}$ &              $7.20\cdot 10^{-3}$ &            $1.42\cdot 10^{-4}$ & $2.07\cdot 10^{-3}$ \\
    KernelNN          &   $6.31\cdot 10^{-5}$ & $1.06\cdot 10^{-2}$ &      $1.34\cdot 10^{-2}$ &              $6.69\cdot 10^{-3}$ &            $1.87\cdot 10^{-4}$ & $5.43\cdot 10^{-3}$ \\
    Point TF &   $4.42\cdot 10^{-5}$ & $1.03\cdot 10^{-2}$ &      $\uwave{7.25\cdot 10^{-3}}$ &              $\uwave{4.90\cdot 10^{-3}}$ &            $1.41\cdot 10^{-4}$ & $2.38\cdot 10^{-3}$ \\
    PointGNN          &   $\uwave{2.82\cdot 10^{-5}}$ & $\uwave{\underline{8.83\cdot 10^{-3}}}$ &      $9.02\cdot 10^{-3}$ &              $6.73\cdot 10^{-3}$ &            $\uwave{\underline{1.36\cdot 10^{-4}}}$ & $\uwave{\underline{1.39\cdot 10^{-3}}}$ \\
    \midrule
    CNN               &   $5.31\cdot 10^{-5}$ & $1.11\cdot 10^{-2}$ &      $4.20\cdot 10^{-3}$ &              $6.70\cdot 10^{-4}$ &            $3.69\cdot 10^{-4}$ & $1.43\cdot 10^{-3}$ \\
    NeuralPDE         &   $\underline{8.24\cdot 10^{-7}}$ & $1.12\cdot 10^{-2}$ &      $3.73\cdot 10^{-3}$ &              $5.37\cdot 10^{-4}$ &            $3.03\cdot 10^{-4}$ & $1.70\cdot 10^{-3}$ \\
    ResNet            &   $2.16\cdot 10^{-6}$ & $1.48\cdot 10^{-2}$  &      $\underline{3.21\cdot 10^{-3}}$ &              $\underline{4.90\cdot 10^{-4}}$ &            $1.57\cdot 10^{-4}$ & $1.46\cdot 10^{-3}$ \\
    \midrule
    Persistence       &   $8.12\cdot 10^{-2}$ & $3.68\cdot 10^{-2}$ &      $1.87\cdot 10^{-1}$ &              $1.42\cdot 10^{-1}$ &            $1.47\cdot 10^{-1}$ & $1.14\cdot 10^{-1}$ \\
    \bottomrule
    \end{tabular}

    \label{tab:results_single}
\end{table}

However, for longer-term predictions, the grid-based models outperform the non-grid models as shown in~\Cref{tab:results_rollout}.
For the grid-based models the underlying spatial structure is fixed and they do not need to additionally learn the dependencies between neighboring measurements.
We hypothesize that because of the simpler spatial dependencies, grid-based models are able to generalize better and thus capture the long term evolution of the system more accurately.

\begin{table}[ht]
    \caption{MSE after 16 prediction steps, *~-~ denotes that the system diverges ($MSE>10$). The best perfoming model for each equation has been \underline{underlined}. Additionally, the best non-grid model has been \uwave{underwaved}. A = Advection, B = Burgers', GD = Gas Dynamics, KS = Kuramoto-Sivashinsky, RD = Reaction-Diffusion, W = Wave}
    \centering
    \begin{tabular}{l@{\hskip 0.25cm}l@{\hskip 0.25cm}l@{\hskip 0.25cm}l@{\hskip 0.25cm}l@{\hskip 0.25cm}l@{\hskip 0.25cm}l}
    \toprule
     model &     A &  B & GD &   KS &   RD & W \\
     \midrule
    FeaSt             &   $1.48\cdot 10^{0}$ & $5.61\cdot 10^{-1}$ &      $8.20\cdot 10^{-1}$ &              $3.74\cdot 10^{0}$ &            $1.30\cdot 10^{-1}$ & $1.61\cdot 10^{0}$ \\
    GAT               &   *  & $8.33\cdot 10^{-1}$ &      $1.21\cdot 10^{0}$ &              $5.69\cdot 10^{0}$ &            $3.86\cdot 10^{0}$ & $2.38\cdot 10^{0}$ \\
    GCN               &   * & $1.31\cdot 10^{1}$ &      $7.21\cdot 10^{0}$ &             * &            * & $7.89\cdot 10^{0}$ \\
    GraphPDE          &   $1.08\cdot 10^{0}$ & $7.30\cdot 10^{-1}$ &      $9.69\cdot 10^{-1}$ &              $2.10\cdot 10^{0}$ &            $8.00\cdot 10^{-2}$ & $\uwave{1.03\cdot 10^{0}}$ \\
    KernelNN          &   $8.97\cdot 10^{-1}$ & $7.27\cdot 10^{-1}$ &      $8.54\cdot 10^{-1}$ &              $\uwave{2.00\cdot 10^{0}}$ &            $6.35\cdot 10^{-2}$ & $1.58\cdot 10^{0}$ \\
    Point TF &   $\uwave{6.17\cdot 10^{-1}}$ & $\uwave{\underline{5.04\cdot 10^{-1}}}$ &      $\uwave{6.43\cdot 10^{-1}}$ &              $2.10\cdot 10^{0}$ &            $\uwave{5.64\cdot 10^{-2}}$ & $1.27\cdot 10^{0}$ \\
    PointGNN          &   $6.61\cdot 10^{-1}$ & $1.04\cdot 10^{0}$ &      $7.59\cdot 10^{-1}$ &              $2.82\cdot 10^{0}$ &            $5.82\cdot 10^{-2}$ & $1.31\cdot 10^{0}$ \\
    \midrule
    CNN               &   $1.61\cdot 10^{-3}$ & $5.55\cdot 10^{-1}$ &      $9.95\cdot 10^{-1}$ &              $1.26\cdot 10^{0}$ &            $1.83\cdot 10^{-2}$ & $5.61\cdot 10^{-1}$ \\
    NeuralPDE         &   $2.70\cdot 10^{-4}$ & $6.60\cdot 10^{-1}$ &      $\underline{4.43\cdot 10^{-1}}$ &              $\underline{1.06\cdot 10^{0}}$ &            $2.24\cdot 10^{-2}$ & $\underline{2.48\cdot 10^{-1}}$ \\
    ResNet            &   $\underline{8.65\cdot 10^{-5}}$ & $1.86\cdot 10^{0}$ &      $4.80\cdot 10^{-1}$ &              $1.07\cdot 10^{0}$ &            $\underline{7.05\cdot 10^{-3}}$ & $2.99\cdot 10^{-1}$ \\
    \midrule
    Persistence       &   $2.39\cdot 10^{0}$ & $6.79\cdot 10^{-1}$ &      $1.46\cdot 10^{0}$ &              $1.90\cdot 10^{0}$ &            $2.76\cdot 10^{-1}$ & $2.61\cdot 10^{0}$ \\
    \bottomrule
    \end{tabular}
    \label{tab:results_rollout}
\end{table}

Interestingly, we found that the models specifically designed to learn solving PDEs, such as KernelNN and GraphPDE, were not as good as the other models when the data was low-resolution as opposed to high-resolution data on which they were originally evaluated. 
This suggests that their underlying assumptions may be too strong to handle such data effectively.

\begin{figure}
     \centering
     \begin{subfigure}[b]{0.49\textwidth}
         \centering
         \includegraphics[width=\textwidth]{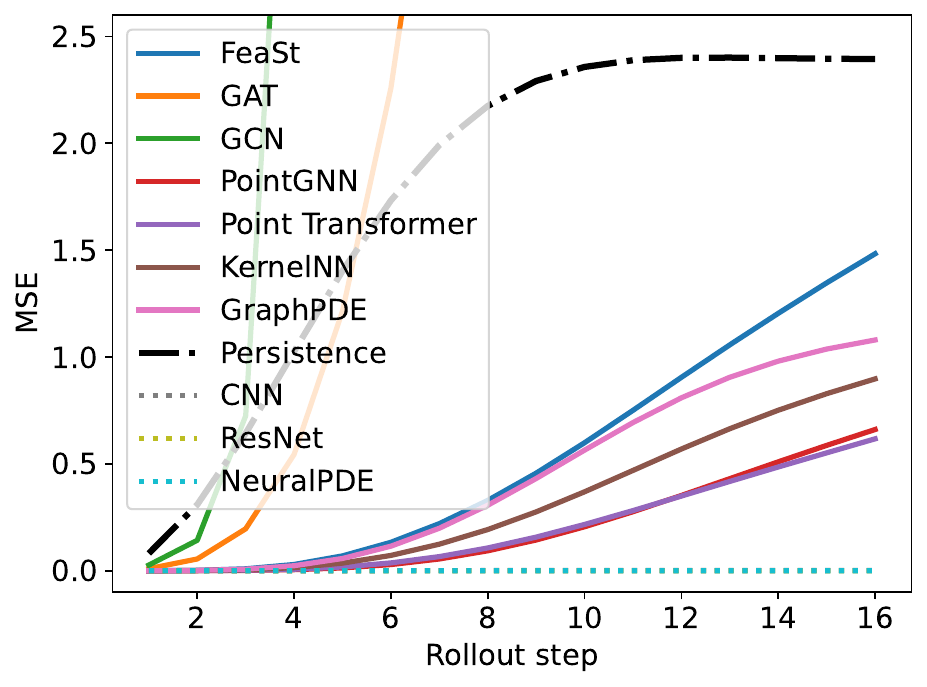}
         \caption{Advection}
         \label{fig:advection_rollout}
     \end{subfigure}
     \hfill
     \begin{subfigure}[b]{0.49\textwidth}
         \centering
         \includegraphics[width=\textwidth]{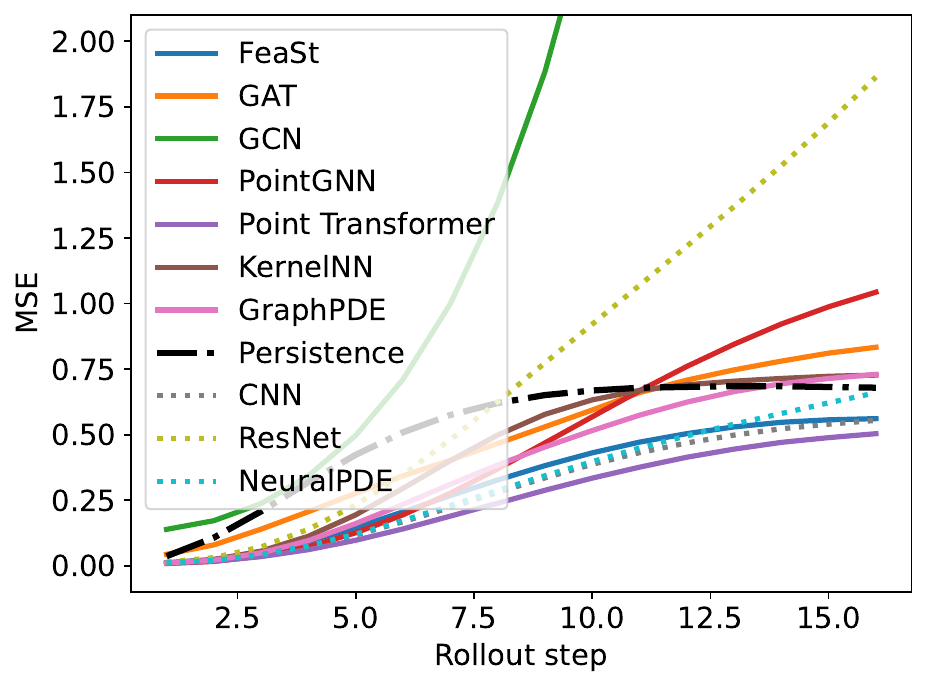}
         \caption{Burgers}
         \label{fig:burgers_rollout}
     \end{subfigure}
     \hfill
     \begin{subfigure}[b]{0.49\textwidth}
         \centering
         \includegraphics[width=\textwidth]{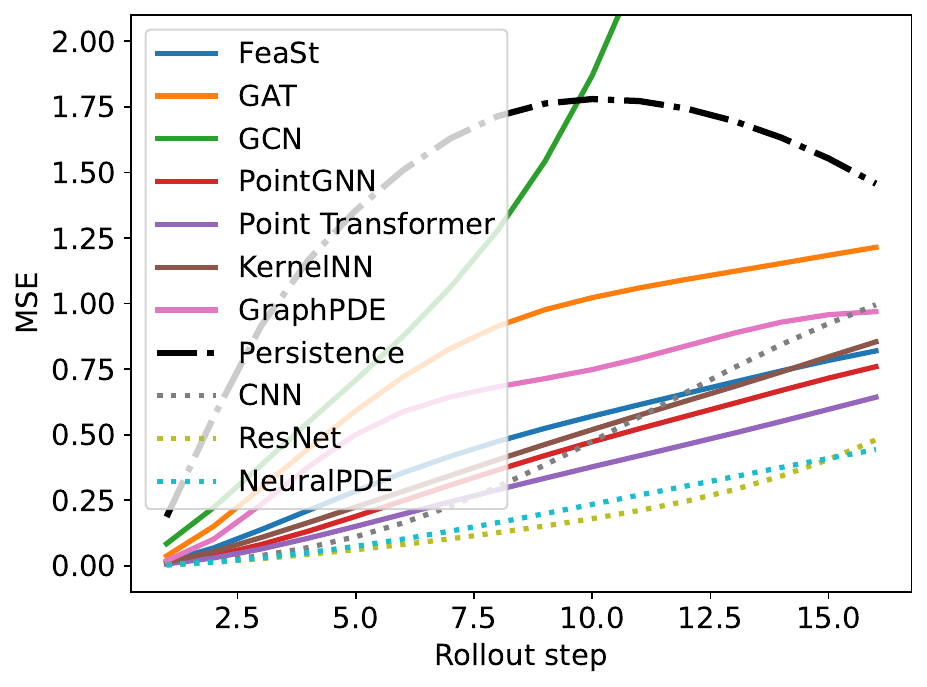}
         \caption{Gas Dynamics}
         \label{fig:gasdynamics_rollout}
     \end{subfigure}
     \hfill
     \begin{subfigure}[b]{0.49\textwidth}
         \centering
         \includegraphics[width=\textwidth]{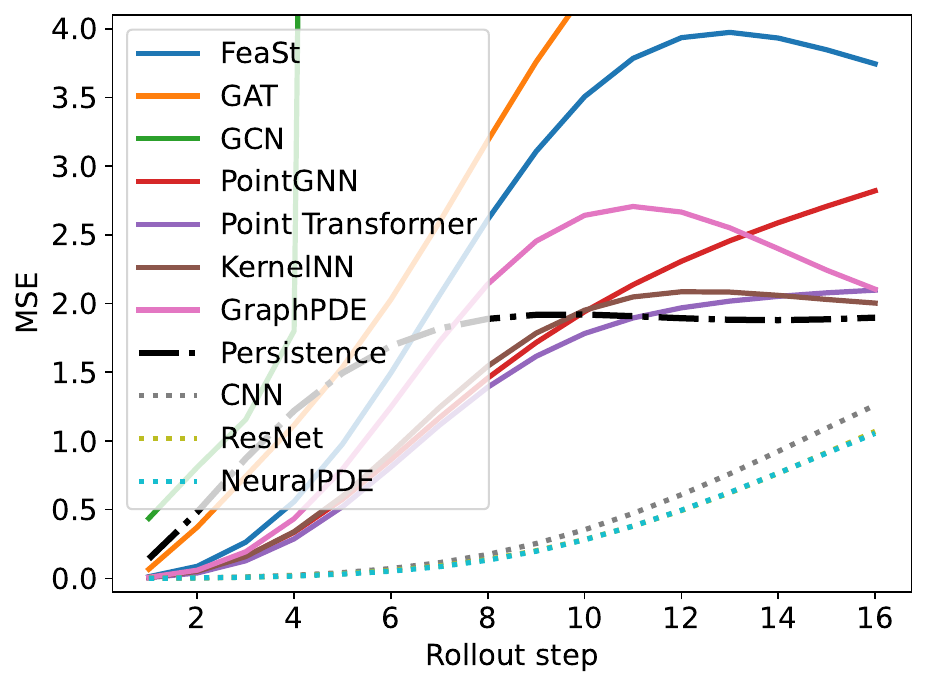}
         \caption{Kuramoto-Sivashinsky}
         \label{fig:kuramotosivashinsky_rollout}
     \end{subfigure}
     \hfill
     \begin{subfigure}[b]{0.49\textwidth}
         \centering
         \includegraphics[width=\textwidth]{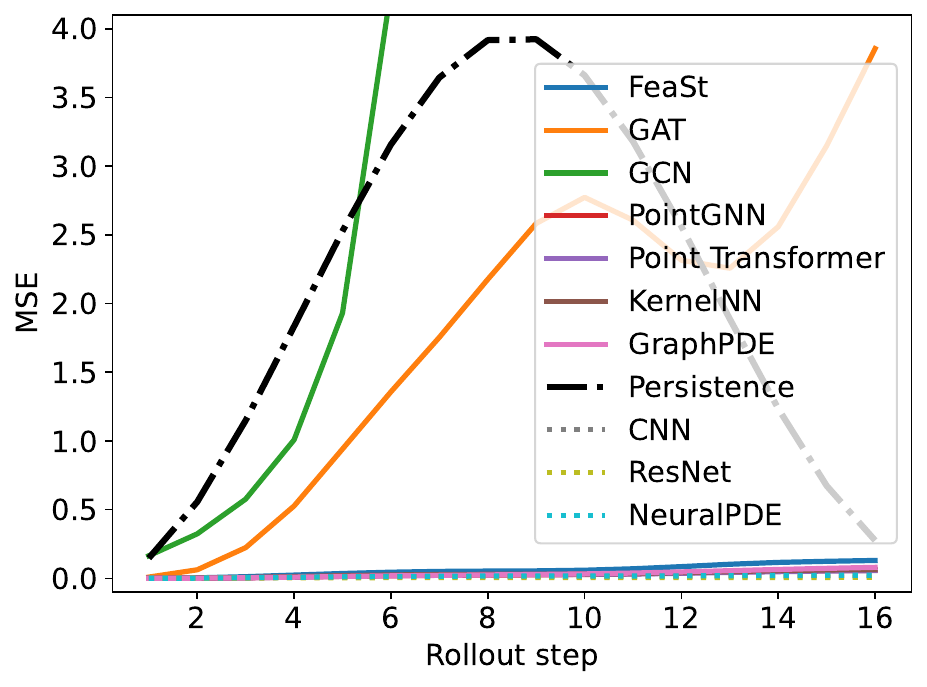}
         \caption{Reaction-Diffusion}
         \label{fig:reactiondiffusion_rollout}
     \end{subfigure}
     \hfill
     \begin{subfigure}[b]{0.49\textwidth}
         \centering
         \includegraphics[width=\textwidth]{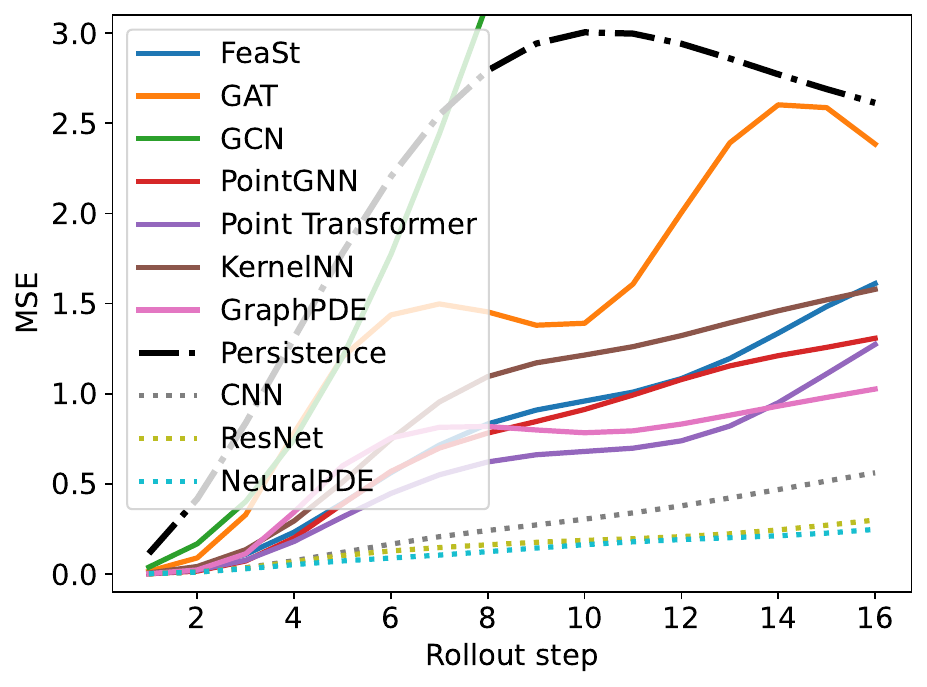}
         \caption{Wave}
         \label{fig:wave_rollout}
     \end{subfigure}
        \caption{Visualization of the accumulation of errors for 16 step predictions for all equations in DynaBench. For better readability, MSEs for diverging predictions are not fully displayed.}
        \label{fig:rollout}
\end{figure}

Additionally, our study brings to light that long-term predictions are still an unsolved challenge for all models. 
The divergence in predictions, as illustrated in~\Cref{fig:rollout}, occurs rapidly and is particularly prominent in systems such as Gas Dynamics and Kuramoto-Sivashinsky equations, where the prediction error exceeds 0.5 after only 16 prediction steps. 
This level of error, which is half of the standard deviation of the data (as explained in~\Cref{sec:postprocessing}), renders it impossible to make use of these long-term predictions. 
Thus, our findings emphasize the need for further research and development in this field to address this issue.

\section{Conclusion}
We have proposed a new benchmark dataset for learning dynamical systems from data under the assumption that measurements are sparse and not structured on a grid. 
This is closer to real-world data than other resources available, as typically measurements are obtained from monitoring stations scattered withing the observation domain.

The DynaBench dataset covers a wide range of physical systems with different properties such as number of connected variables, degree of the differential operators etc. 
We have thoroughly evaluated several models capable of learning functions on arbitrary geometries on the DynaBench dataset, including graph neural networks, point-cloud processing models and several state-of-the-art approaches.
Our results show that the selected models are on par with state-of-the-art grid models in providing accurate short-term predictions, but long-term forecasting remains an open challenge. 

We hope that the release of DynaBench will facilitate and encourage research in this area, leading to advancements in the state-of-the-art and as a consequence more accurate models for real-world data, which our benchmark is mirroring.

\clearpage

\section*{Ethical statement}
This research paper proposes a benchmark dataset and evaluates several machine learning models for learning dynamical systems from data. The use of benchmarking is a common practice in the machine learning community to compare different models in a standardized setting. Synthetic datasets are used because they allow for a controlled environment and can be generated easily. However, it should be noted that synthetic data can never perfectly represent real-world data, and as such, every model should also be evaluated on real-world data before being used in critical applications.

Potential risks associated with incorrect predictions of important systems such as weather and climate simulations or electromagnetic field simulations for safety assessment should be discussed thoroughly.
Synthetic datasets can provide a useful starting point for model evaluation and the development of new approaches, but they need to be assessed on domain-specific data for real-world deployment. 
Particularly for safety-critical applications.
While our proposed benchmark dataset and evaluated machine learning models provide useful insights into learning dynamical systems, they should not be used as the sole basis for making important political decisions, particularly concerning weather or climate data.

 While data-driven approaches have again and again shown their superiority over classical methods in a variety of applications, they are also prone to overfitting and adversarial attacks, if not carefully designed and validated.
 The risks and benefits of replacing existing numerical simulations or expert knowledge with deep learning approaches should always be taken into account and thoroughly discussed when developing and applying new models.
 Any decision based on machine learning models should be made after considering the potential sources of errors the models introduce, as well as the lack of explainability of black-box approaches.



\bibliography{references.bib}
\bibliographystyle{splncs04.bst}

\end{document}